\documentclass[conference,10pt]{IEEEtran}
\usepackage[T1]{fontenc}

\usepackage{amssymb,amsmath,amsfonts,amsthm}
\usepackage{mathrsfs}
\usepackage{cite}
\usepackage{array}
\usepackage{tabularx}
\usepackage[table]{xcolor}
\usepackage{multicol, multirow}
\usepackage{color}
\usepackage{mathtools}
\usepackage{subcaption}
\usepackage{mathtools}
\usepackage{tikz,pgf}
\usetikzlibrary{calc,positioning,mindmap,trees,decorations.pathreplacing}
\usepackage{graphicx}
\usepackage{bbm}
\usepackage[utf8]{inputenc}
\usepackage{algorithm}

\IEEEoverridecommandlockouts
\ifCLASSINFOpdf
\else
\fi
\usepackage[cmintegrals]{newtxmath}
\begin{document}
	\title{Device Scheduling and Update Aggregation Policies for Asynchronous Federated Learning}
	
	\author{Chung-Hsuan Hu, Zheng Chen, and Erik G. Larsson\\
		Department of Electrical Engineering, Link\"{o}ping University, Sweden.\\ Email: \{chung-hsuan.hu, zheng.chen, erik.g.larsson\}@liu.se
	\thanks{This work was supported in part by Centrum för Industriell Informationsteknologi (CENIIT), Excellence Center at Link\"{o}ping - Lund in Information Technology (ELLIIT), and Knut and Alice Wallenberg (KAW) Foundation.}}
	
	\maketitle
	
\begin{abstract}	
Federated Learning (FL) is a newly emerged decentralized machine learning (ML) framework that combines on-device local training with server-based model synchronization to train a centralized ML model over distributed nodes. In this paper, we propose an asynchronous FL framework with periodic aggregation to eliminate the straggler issue in FL systems. For the proposed model, we investigate several device scheduling and update aggregation policies and compare their performances when the devices have heterogeneous computation capabilities and training data distributions. From the simulation results, we conclude that the scheduling and aggregation design for asynchronous FL can be rather different from the synchronous case. For example, a norm-based significance-aware scheduling policy might not be efficient in an asynchronous FL setting, and an appropriate ``age-aware'' weighting design for the model aggregation can greatly improve the learning performance of such systems.
\end{abstract} 


\section{Introduction}
Federated learning (FL) is a decentralized machine learning (ML) framework with multiple devices collaboratively participating in a common training process over locally distributed data \cite{konevcny2016federated}. 
In contrast to centralized ML where the entire training data are stored in a central unit, in an FL system, the local training data can be kept private in each device without being uploaded to a cloud/server. Compared to prior distributed optimization frameworks that assume evenly distributed data, FL considers more practical settings where the devices might be massively distributed, with non-IID and unbalanced data.

One common problem in FL systems using the original Federated Averaging (FedAvg) algorithm proposed in \cite{mcmahan2017communication} and its different variations is the straggler issue. This problem originates from the fact that, due to synchronized training and updating, the time duration of one communication round is strictly limited by the slowest participating device \cite{chen2016revisiting}. In a practical environment with heterogeneous devices in terms of their computation capabilities, the straggler issue has a significant impact on the completion time of an FL process.
One possible solution to tackle this problem is to shift from the synchronous setting in FedAvg to asynchronous training and updating, to avoid waiting for straggling devices before update aggregation. Several deep learning algorithms with asynchronous FL \cite{zhang2015staleness, xie2019asynchronous} have been studied in the literature. Moreover, various heuristic aggregation policies have been proposed to deal with the increased variation of local updates caused by the asynchronous structure \cite{chen2020asynchronous, chai2020fedat}. 

In a wireless FL system, the uploading of local updates takes place over the wireless uplink, which will be the part that is most affected by the scarcity of wireless resources. To reduce the communication load in this procedure, one solution is to allow only a fraction of participating devices to upload their local updates in each communication round. Device scheduling and resource allocation have therefore become an important design aspect for FL over wireless networks \cite{gafni2021federated, Yang2020scheduling}.  
In traditional cellular networks, the purpose of device scheduling is usually associated with maximizing spectral efficiency or network throughput. 
However, for distributed learning systems such as FL, the system objective is to optimize the parameters in the training model.
Device scheduling for FL requires learning-oriented instead of rate-oriented design, which makes this problem fundamentally different from the existing solutions in conventional cellular networks.
Intuitively, a device with a higher potential impact on the learning-related system performance should be given higher priority to be scheduled, and possibly a larger amount of communication resources to convey their information.
Several existing works consider different metrics to indicate the significance of local updates, such as norm of the model updates \cite{amiriconvergence}, signal-to-noise-ratio and data uncertainty \cite{importance-aware}, success probability of update transmission \cite{salehi2020federated}, and Age-of-Update (AoU) \cite{yang2019agebased}. However, all of them consider synchronous FL based on the original form of the FedAvg algorithm. Few existing works have considered device scheduling and resource allocation for asynchronous FL \cite{Lee2021adaptive}, which makes it a highly under-explored area. 

The main purpose of this work is to answer the following questions:
\begin{itemize}
	\item For asynchronous FL with heterogeneous devices, what is the most appropriate scheduling policy given limited communication resources? 
	\item Under a certain scheduling policy, how should we design the update aggregation rule?
	\item What are the fundamental differences between synchronous and asynchronous FL systems in terms of the joint design of scheduling and aggregation policy?   
\end{itemize}

To answer these questions, we investigate several schemes for device scheduling and model update aggregation in asynchronous FL systems under device heterogeneity in their computation capacity and training data distribution.\footnote{Analytical evaluation on convergence of the proposed schemes will be addressed in an extended version of this paper.} The performance of the proposed schemes are evaluated and compared based on a classification problem using the MNIST data set \cite{lecun-mnisthandwrittendigit-2010}.

\section{System Model}

\label{sec:system-model}

We consider an FL system with $N$ devices participating in training a shared global learning model, parameterized by a $d$-dimensional parameter vector $\boldsymbol{\theta}\in\mathbb{R}^d$. Denote $\mathcal{N}=\{1,...,N\}$ as the set of device indices in the system. Each device $k\in\mathcal{N}$ holds a set of local training data $\mathcal{S}_k$ with size $|\mathcal{S}_k|$. Let $\mathcal{S}=\cup_{k\in\mathcal{N}}\mathcal{S}_k$ represent the entire data set in the system with size $|\mathcal{S}|$, where $\mathcal{S}_i\cap \mathcal{S}_j=\emptyset$, $\forall i\neq j$.  The objective of the system is to find the optimal parameter vector $\boldsymbol{\theta}^*$ that minimizes an empirical loss function defined by
\begin{equation}
	F(\boldsymbol{\theta})=\frac{1}{|\mathcal{S}|}\sum_{x\in\mathcal{S}}l(\boldsymbol{\theta},x),
	\label{eq:globalLoss}
\end{equation}
where $l(\boldsymbol{\theta},x)$ is the sample-wise loss function computed over the data sample $x$. We define the local loss function at device $k$ as 
\begin{equation}
	F_k(\boldsymbol{\theta})=\frac{1}{|\mathcal{S}_k|}\sum_{x\in\mathcal{S}_k}l(\boldsymbol{\theta},x),
	\label{eq:localLoss}
\end{equation}
which is averaged over the local training data set.
Then, we can rewrite $\eqref{eq:globalLoss}$ as
\begin{equation}
	F(\boldsymbol{\theta})=\sum_{k\in\mathcal{N}}\frac{|\mathcal{S}_k|}{|\mathcal{S}|}F_k(\boldsymbol{\theta}).
\label{eq:globalLoss_f}
\end{equation}     
	
\subsection{FedAvg with Synchronous Training and Aggregation}

\begin{figure}[t!]
	\centering
	\includegraphics[width=\columnwidth,height=3.8cm]{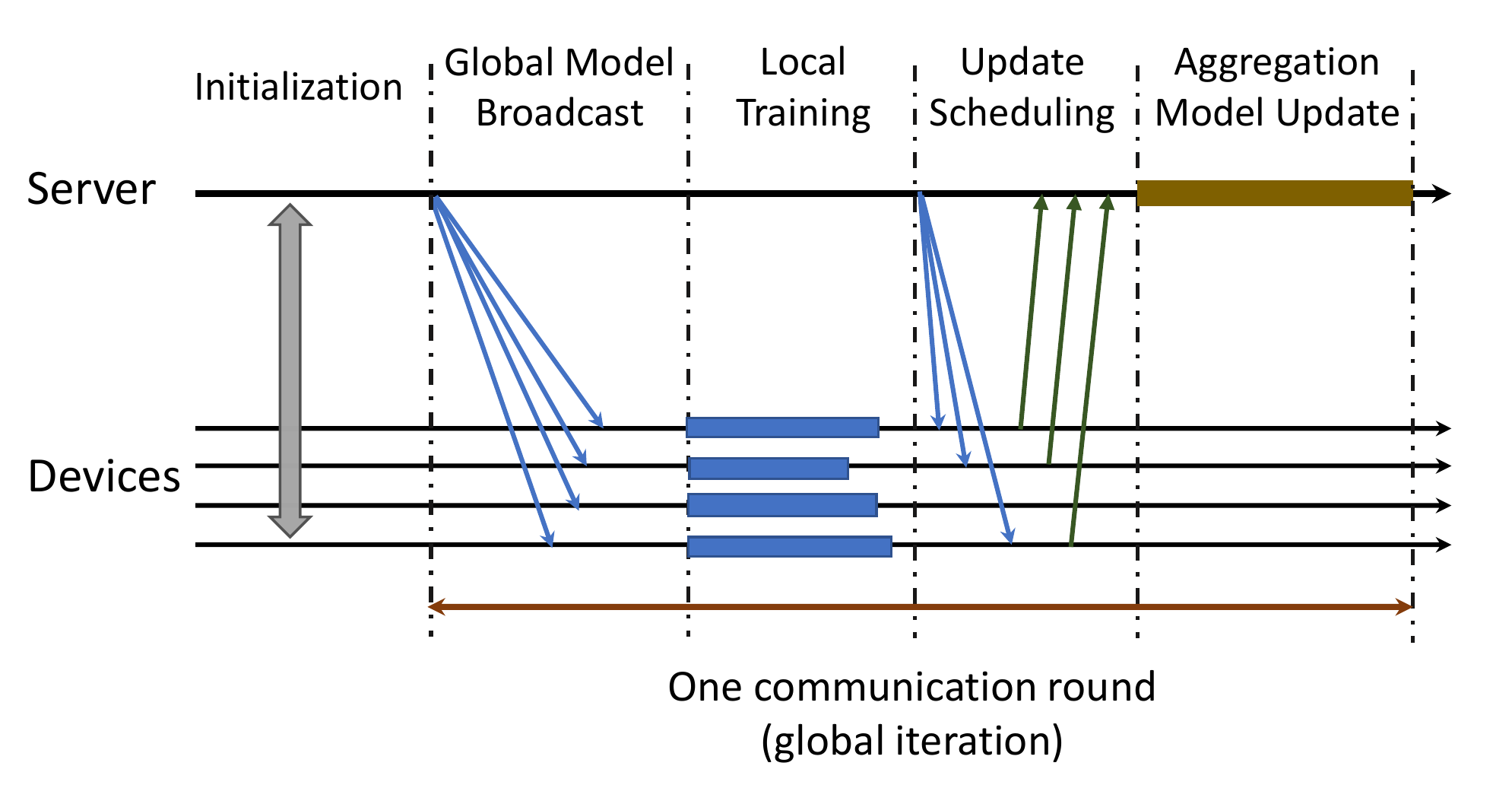}
	\caption{The FL process and information exchange between the server and the participating devices.}
	\label{fig:FL}
	\vspace{-.2cm}
\end{figure}
In a typical FL system, all the devices participate in the global training process following a synchronized procedure. FedAvg\cite{mcmahan2017communication} is widely considered as a representative scheme with this synchronous structure of local training and global aggregation. In FedAvg algorithm, the entire training process is divided into many global iterations (communication rounds), where during every global iteration, the server aggregates the received stochastic gradient updates from the participating devices, computed over their locally available data. 

We consider a modified version of the FedAvg algorithm, where an extra step of device scheduling is added after local training, as illustrated in Fig.~\ref{fig:FL}. The main motivation behind this consideration is to reduce the communication costs and delay, especially in a wireless network with limited data rates. In the $t$-th global iteration with $t=1,2,\ldots$, the following steps are executed:
\begin{enumerate}
	\item The server broadcasts the current global model $\boldsymbol{\theta}(t)$ to the device set $\mathcal{N}$.
	\item Each device $k$ runs $E$ times of stochastic gradient descent (SGD) iteration and the update rule follows
	\begin{equation}
		\boldsymbol{\theta}_{k}(t,\tau+1)=\boldsymbol{\theta}_{k}(t,\tau)-\alpha_{k}(t,\tau)\nabla F_k(\boldsymbol{\theta}_{k}(t,\tau),\mathcal{B}_k(t,\tau)),
		\label{eq:localItUpdate}
	\end{equation}
	with $\tau=0,...,E-1$ being the local iteration index, $\boldsymbol{\theta}_{k}(t,0)=\boldsymbol{\theta}(t)$, $\alpha_{k}(t,\tau)$ representing the learning rate and $\nabla F_k(\boldsymbol{\theta}_{k}(t,\tau),\mathcal{B}_k(t,\tau))$ being the gradient computed based on a randomly selected mini-batch $\mathcal{B}_k(t,\tau)\subseteq\mathcal{S}_k$.
    After completing the local training, $\boldsymbol{\theta}_{k}(t,E)$ is the local update from device $k$.
	\item Due to limited wireless resources, only a subset of devices $\Pi(t)\subseteq\mathcal{N}$ is eligible for uploading their local model updates to the server. Similar consideration can be found in \cite{amiriconvergence} and \cite{importance-aware}. Note that such update scheduling is not considered in FedAvg, i.e. $\Pi(t)=\mathcal{N}$.
	\item After receiving the local updates from the scheduled devices, the server aggregates the received information and updates the global model as
	\begin{equation}
		\boldsymbol{\theta}(t+1)=\sum_{k\in\Pi(t)}\frac{|\mathcal{S}_k|}{|\mathcal{S}|}(\boldsymbol{\theta}_{k}(t,E)-\boldsymbol{\theta}(t))+\boldsymbol{\theta}(t),
		\label{eq:syncFlAggregation}	
	\end{equation}
 where the update from each non-scheduled device $k$ is $\boldsymbol{\theta}_{k}(t,E)-\boldsymbol{\theta}(t)=0$, thus is omitted in \eqref{eq:syncFlAggregation}.
\end{enumerate}
Such iterative procedure continues until the system converges.
\label{sec:fedavg}  

\subsection{Asynchronous FL with Periodic Aggregation}

\begin{figure}
	\centering
	\includegraphics[width=\columnwidth,height=3.5cm]{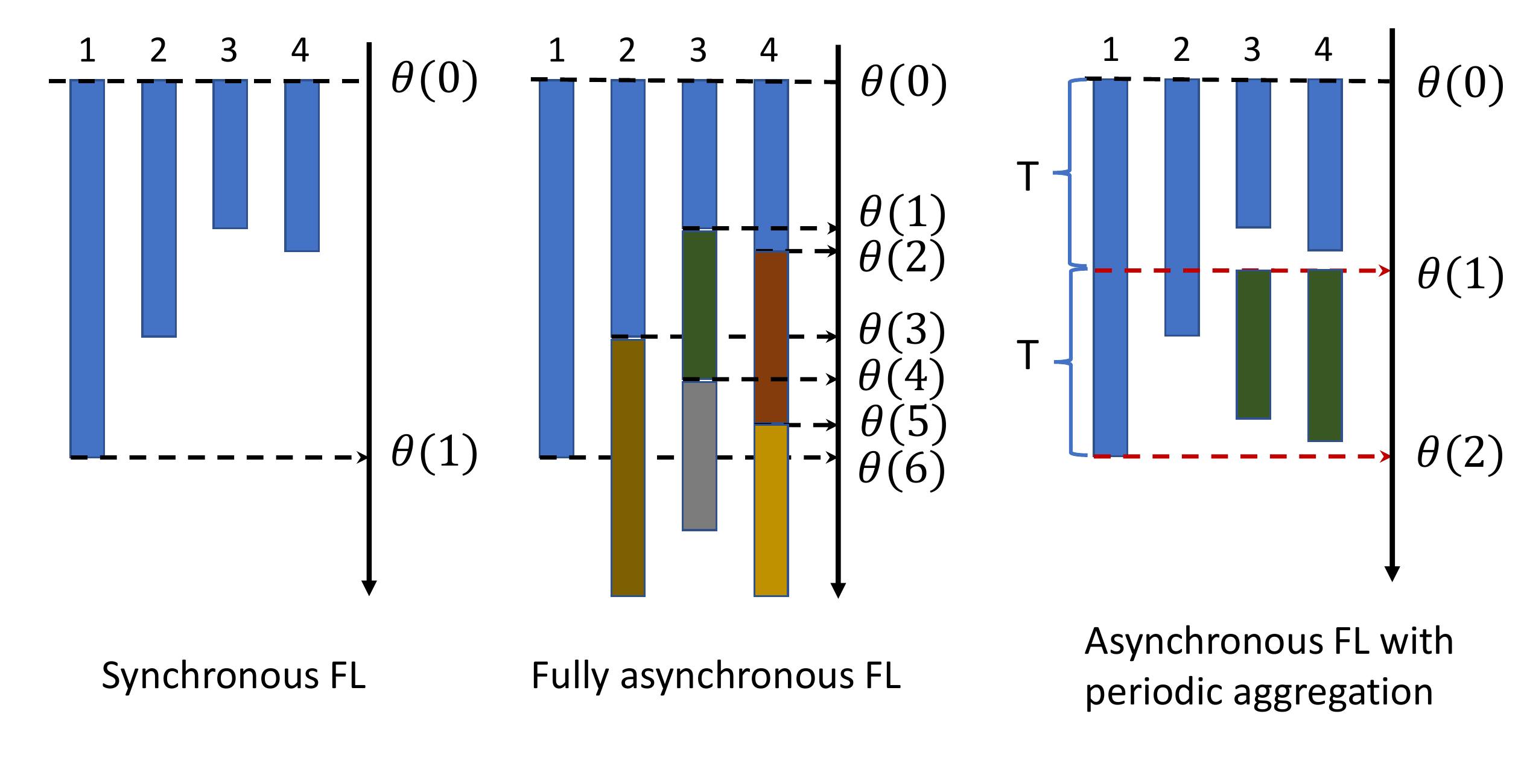}
	\vspace{-.5cm}
	\caption{Illustration of the conceptional differences between synchronous FL, fully asynchronous FL in \cite{xie2019asynchronous}, and our proposed asynchronous FL with periodic aggregation. $\boldsymbol{\theta}(t)$ represents the model parameter in the $t$-th global iteration.}
	\label{fig:FL-syn-asyn}
	\vspace{-.2cm}
\end{figure}

In an FL system with synchronous training and updating, the update aggregation is feasible only after all the involved devices finish their local training (SGD computation) step. This implies those with inferior computation capability introduce the straggler issue and slow the training process. To address the issue, asynchronous FL has been proposed in \cite{xie2019asynchronous, chen2020asynchronous}, which proves the effectiveness of resolving the issue. However, fully asynchronous FL with sequential updating can face the problem of high communication costs caused by frequent model updating and transmission of local updates. To tackle the aforementioned concerns, we propose an asynchronous FL framework with periodic aggregation. The general idea is to allow asynchronous training at different devices, with the server periodically collecting updates from those devices that have completed their computation, while the rest continue their local training without being interrupted or dropped. Fig.~$\ref{fig:FL-syn-asyn}$ shows an example of the training and updating timeline of the original synchronous FL, fully asynchronous FL \cite{xie2019asynchronous}, and our proposed scheme.    

We consider the scenario where the participating devices have different computation capabilities. After each device finishes its local training, it sends a signal to the server indicating its readiness for update reporting. After every time duration $T$, the server schedules a subset of ready-to-update devices. The received local updates will be aggregated at the server by applying some weighted averaging rule. The updated global model will be again distributed to all the ready devices, which will then continue their local SGD steps based on the newly received global model. The key design factors in this particular asynchronous FL setup lie in two folds:
\begin{enumerate}
	\item At each aggregation time, the set of ready-to-update devices might be different. Given the communication resource constraint, how should we schedule the subset of available devices for update reporting?
	\item The updated local models from different devices might be obtained from different previously received global models, some are more recent and some are more outdated. How should we design an appropriate aggregation and weighting policy taking into account the freshness of model updates in this asynchronous setting?   
\end{enumerate}

In the remainder of this paper, we will investigate several scheduling and aggregation policies for FL with asynchronous training and periodic data aggregation. We define $\mathcal{K}(t)$ as the set of all ready-to-update devices in the $t$-th global iteration. Let $\Pi(t)\subseteq \mathcal{K}(t)$ be the set of scheduled devices, with $|\Pi(t)|=\min\{R,|\mathcal{K}(t)|\}\triangleq R'(t)$, indicating up to $R$ devices are scheduled.\footnote{We consider a simplified system model with limited communication budget and uniform gradient update precision for the scheduled devices. More realistic channel models and corresponding compression scheme will be further investigated in extended studies.} For any device $k\in \Pi(t)$, its local update is computed according to \eqref{eq:localItUpdate}, with the initial model parameter vector in the first local iteration being
\begin{equation}
	\boldsymbol{\theta}_{k}(t,0)=\boldsymbol{\theta}(s_k(t)),
\end{equation}
where
\begin{equation}
	s_k(t)=\underset{t'<t}{\mbox{max}}\{t'|k\in \mathcal{K}(t')\}+1
\end{equation} 
specifically indicates the latest global iteration index of which device $k$ has received an updated global model. We adopt a regularization technique proposed in \cite{chai2020fedat} to alleviate potential model imbalance caused by asynchronous training. In \eqref{eq:localItUpdate}, $\nabla F_k(\boldsymbol{\theta}_{k}(t,\tau),\mathcal{B}_k(t,\tau))$ is computed based on the following regularized local loss
\begin{equation}
	\frac{1}{|\mathcal{B}_k|}\sum_{x\in\mathcal{B}_k}l(\boldsymbol{\theta}_{k}(t,\tau),x)+\frac{\lambda}{2}\|\boldsymbol{\theta}_{k}(t,\tau)-\boldsymbol{\theta}_{k}(t,0)\|_2^2,
\end{equation}
where $l(\boldsymbol{\theta}_{k}(t,\tau),x)$ is introduced in \eqref{eq:globalLoss} and $\lambda>0$ is the regularization coefficient. 
Inspired by the concept of Age of Information (AoI) \cite{kosta2017age}, we define a metric "Age of Local Update" (ALU) as
\begin{equation}
	a_k(t)=t-s_k(t),
\end{equation}
which shows the elapsed time since the last reception of an updated global model.\footnote{Note that this age-based definition is different from the Age of Update (AoU) proposed in \cite{yang2019agebased}, which measures the elapsed time at each device since its last participation in model aggregation.} After receiving the scheduled updates, the model aggregation at the server is conducted as
\begin{equation}
	\boldsymbol{\theta}(t+1)=\sum_{k\in\Pi(t)}w_k(t)\boldsymbol{\theta}_{k}(t,E)+\sum_{k\in\mathcal{N}\backslash\Pi(t)}w_k(t)\boldsymbol{\theta}_{k}(t,0),	
	\label{eq:update}
\end{equation}
with $\sum_{k\in\mathcal{N}}w_k(t)=1$.  Here, each weight coefficient $w_k(t)$ can be related to the training data size $|\mathcal{S}_k|$, or the ALU $a_k(t)$, or the combination of both. Two different weighting designs for non-scheduled devices may be considered, depending on the assumption on the statistical distribution of local updates: for any non-scheduled device $k\in\mathcal{N}\backslash\Pi(t)$:
\begin{enumerate}
	\item $w_k(t)=0$. This design inherently assumes that all devices in $\mathcal{N}$ have identical training data distribution such that averaging over $\Pi(t)$ is statistically equal to averaging over $\mathcal{N}$. Note that this approach has been considered in \cite{amiriconvergence}, \cite{li2019convergence}.
	\item $w_k(t)\neq0$. This design considers the non-updated local models from the set of non-scheduled devices in the aggregation step, as in \cite{mcmahan2017communication}. 
\end{enumerate}	
\label{sec:ayncFL}

\section{Scheduling and Aggregation Policies for Asynchronous FL}
As mentioned in Section~\ref{sec:system-model}, in every global iteration, only a subset of all available devices $\Pi(t)\subseteq\mathcal{K}(t)$ will be scheduled for uploading their model updates. Among the scheduled devices in $\Pi(t)$, their local updates might have different levels of data freshness, since their local gradients are computed based on different previously received global models. Moreover, the aggregation-involving rate of each device might be different, as a result of device scheduling and asynchronous training, which places an unbalanced contribution to the global model. These issues suggest a joint consideration of device scheduling and aggregation policy to leverage data freshness and significance in an asynchronous FL setting.

\subsection{Device Scheduling Policies}
We consider three different scheduling policies, namely random, significance-based and frequency-based scheduling, which are described as follows.
	\subsubsection{Random Scheduling}
	We select $R'(t)$ devices randomly from the set of ready-to-update devices without replacement. This often serves as a baseline policy in the literature of FL.
	\subsubsection{Significance-based Scheduling}
	Under the preference of devices with more influential updates, we first sort the norm of gradient update from all available devices,
	\begin{equation}
		\|\boldsymbol{\theta}_{k}(t,E)-\boldsymbol{\theta}_{k}(t,0)\|_2,
		\label{eq:norm}
	\end{equation} 
    and then select those with $R'(t)$ largest values. Note that this approach requires the quantity in \eqref{eq:norm} to be shared as side information to the server. 
	\subsubsection{Frequency-based Scheduling}
	To maintain the balance among the devices in terms of their contribution to the global model, this policy assigns higher preference to devices with lower aggregation-involving rate during previous communication rounds. To explain the idea, we define a counting metric
	\begin{equation}
		c_k(t)=
		\begin{cases}
		  0, & t=0 \\
		  \sum_{t'=0}^{t-1}\boldsymbol{1}(k\in\Pi(t')), & \text{otherwise}
        \end{cases}
	\end{equation}
	which characterizes how many times a device has been previously scheduled for uploading its local updates.
	After sorting $c_k(t)$ of all available devices, those with the $R'(t)$ smallest values are selected. If multiple devices have the same counting metric value, random selection will be performed accordingly.
	
	\subsection{Update Aggregation Policies}
	To address the model aggregation with asynchronous updates and various data proportions, we consider two types of aggregation policies, namely equal weight and age-aware aggregation.
	\subsubsection{Equal Weight}
	With this policy, the model updates from all selected devices are aggregated with uniform weighting, which means that the weights are assigned purely based on their data proportion, regardless of their last received global model. Hence, the aggregation weight of device $k$ in the $t$-th global iteration is determined by
	\begin{equation}
	   w_k(t)=\frac{|\mathcal{S}_k|}{\sum_{i\in\mathcal{P}}|\mathcal{S}_i|}\boldsymbol{1}(k\in\mathcal{P}), \forall k\in\mathcal{N}
    \end{equation}
    where $\mathcal{P}=\Pi(t)$ or $\mathcal{P}=\mathcal{N}$, depending on the assumption on the data distributions of the non-scheduled devices, as discussed in Section~\ref{sec:ayncFL}. Note that this is often considered as a baseline aggregation rule for FL.

\subsubsection{Age-aware Aggregation}
	We consider two options for the age-aware aggregation policy. The first option is to assign a higher weight to the older local updates, i.e., those with larger $a_k(t)$. This choice might balance the participation rate among different devices and potentially reduce the risk of model training excessively biased to those with stronger computation capacity. However, it also creates the doubt of applying outdated updates on an already evolved global model. On the contrary, the second option is to assign higher weights to those with smaller $a_k(t)$. By favoring fresher local updates it might help the global model to converge smoothly with time, at the risk of converging to an imbalanced model, especially in the scenario with non-IID data distribution.

	The age-aware weighting design is given by
	\begin{equation}
		w_k(t)=\frac{|\mathcal{S}_k|{\gamma}^{a_k(t)}}{\sum_{i\in\mathcal{P}}|\mathcal{S}_i|{\gamma}^{a_i(t)}}\boldsymbol{1}(k\in\mathcal{P}), \forall k\in\mathcal{N},
		\label{eq:wk_t_age}
	\end{equation}
    where $\mathcal{P}=\Pi(t)$ or $\mathcal{P}=\mathcal{N}$. Here, $\gamma$ is a real-valued constant factor, which can be divided into two cases:
    \begin{itemize}
    	\item $\gamma>1$, the system favors older local updates.
    	\item $\gamma<1$, the system favors fresher local updates.
    \end{itemize}
	
\section{Simulation Results}
In this section, we evaluate the performance of the combination of different scheduling and aggregation policies using the MNIST training data set for the hand-written digit classification problem \cite{lecun-mnisthandwrittendigit-2010}. The data set has $|\mathcal{S}|=60000$ samples that are allocated evenly to $N=100$ devices. The general system setting is described as follows.
	\begin{itemize}
		\item The local training duration of each device $k$ in every global iteration, denoted by $T_k$, is generated by a uniform distribution $T_k\sim\mathcal{U}(0,T_{\max})$, where $T_{\max}$ represents the longest computing time due to straggling issue.
		\item For the asynchronous FL scheme, we choose $T=T_{\max}/4$ as the aggregation period.
		\item In the case with IID data distribution, each device possesses an equal amount of disjoint samples $(\boldsymbol{x},y)$ randomly picked from $\mathcal{S}$. Under the non-IID setting, the data allocation is determined by using the same method as in \cite{mcmahan2017communication}. Under this setting, each device contains data samples of at most two different digits.
		\item In every global iteration, up to $R=30$ devices are scheduled for uploading their local model updates.
		\item The learning rate is $\alpha_k(t,\tau)=0.01, t\leq20$ and $\alpha_k(t,\tau)=0.005, 20<t\leq40$, $\forall \tau, k$.
		\item The regularization coefficient is $\lambda=0.02$. 
	\end{itemize}
In the aggregation process, we consider the case with $w_k(t)=0$ for $k\in\mathcal{N}\backslash\Pi(t)$, meaning that the updated global model is purely computed based on the received updates from the scheduled devices, as explained in Section \ref{sec:ayncFL}. The implementation of FedAvg is also modified accordingly with the same weighting design.

\begin{figure}[t!]
	\centering
	\includegraphics[width=\columnwidth,height=5.8cm]{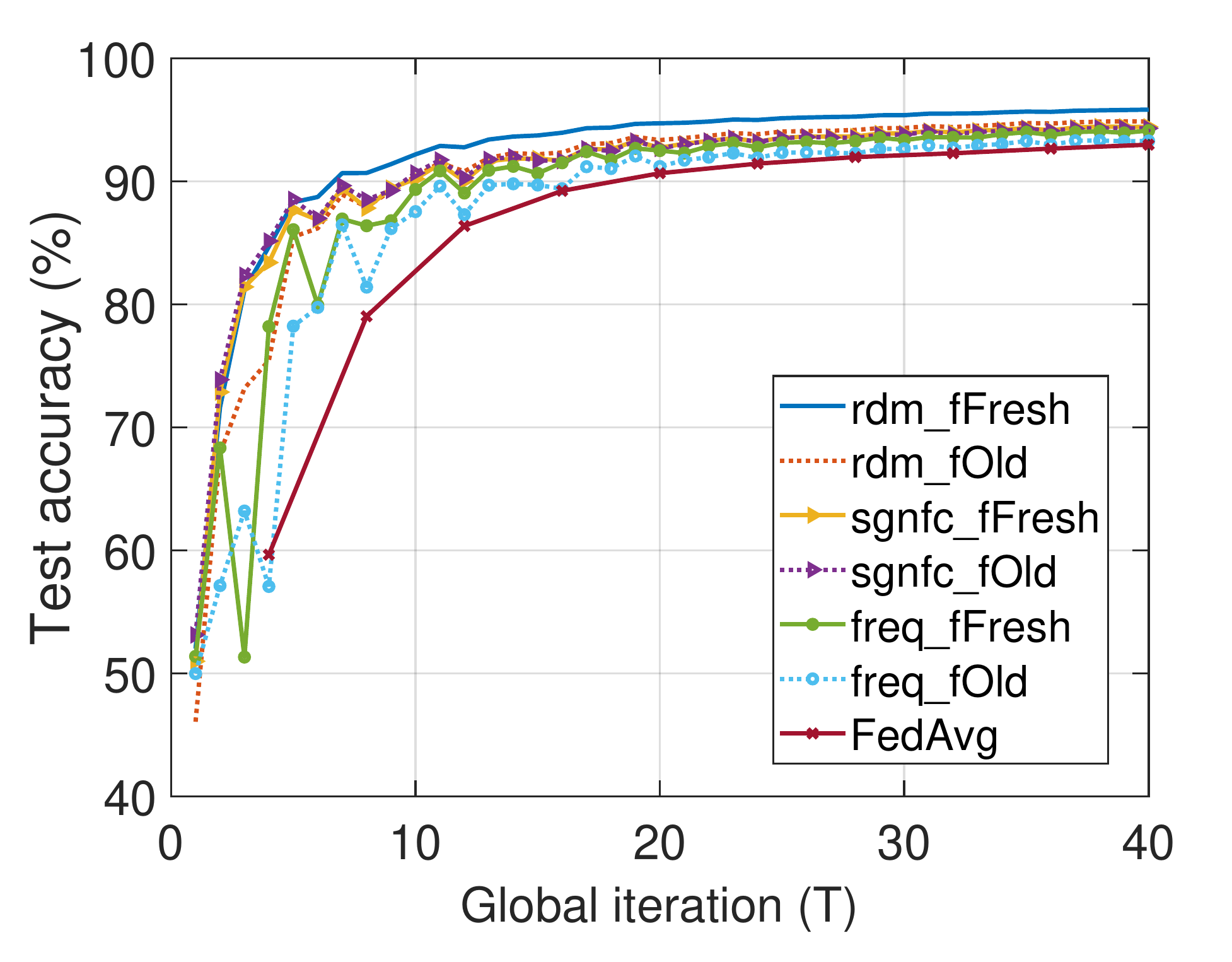}
		\caption{Test accuracy under IID data distribution.}
	\label{fig:algo-100}
	\hfill
	\vspace{-.6cm}
\end{figure}

\begin{figure}[t!]
	\centering
	\includegraphics[width=\columnwidth,height=5.8cm]{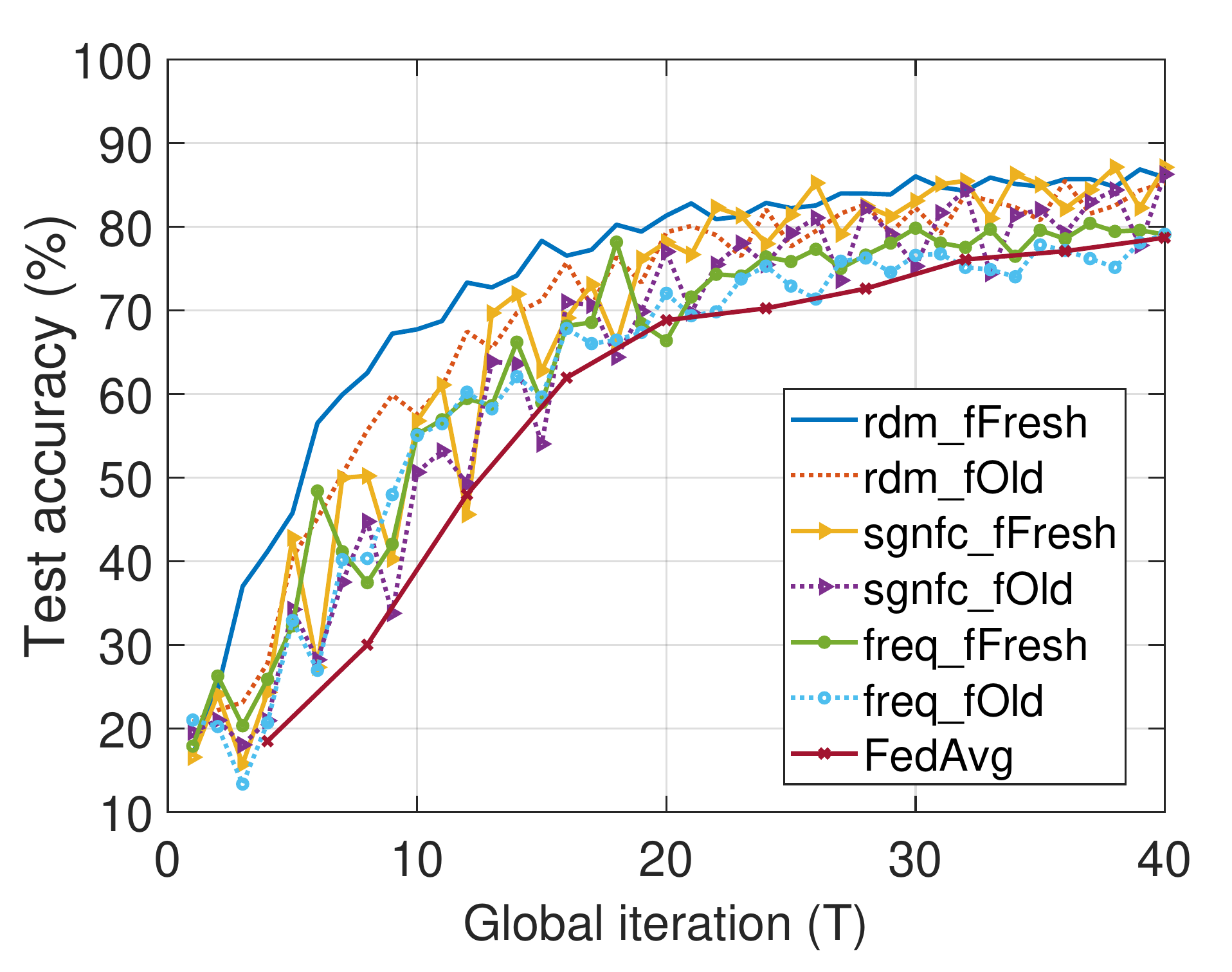}
	\caption{Test accuracy under non-IID data distribution.}
	\label{fig:algo-50}
	\hfill
	\vspace{-.5cm}
\end{figure}

Figs. \ref{fig:algo-100} and \ref{fig:algo-50} show the test accuracy comparison between different scheduling and aggregation policies for asynchronous FL under IID and non-IID data distributions, respectively. The performance of FedAvg is also presented as a baseline scheme. Since the aggregation period in the asynchronous FL case is $T=T_{\max}/4$, the model aggregation in asynchronous FL is four times more frequent than in the case with FedAvg. The abbreviations in the legends are summarized in Table \ref{tab:legend}.
	
From the simulation results, we first observe that the asynchronous FL scheme generally outperforms FedAvg in both IID and non-IID data scenarios. However, this advantage comes at the cost of more frequent local update uploading and aggregation, which leads to higher communication costs. Besides, among the considered scheduling and aggregation policies for asynchronous FL, the combination of random scheduling and age-aware aggregation favoring fresher local updates shows superior performance compared to the others. We have observed that with other data distribution scenarios, especially when slower devices possess some unique training data, favoring older updates sometimes performs better. Particularly, we observe that significance-based scheduling leads to fluctuating test performance due to the increased variation in the model aggregation. This suggests that for asynchronous FL, norm-based significance-aware update scheduling might not be an appropriate option. Another observation is that, among all the scheduling policies, frequency-based scheme has generally the worst performance, which shows that imposing equal participation rate among the agents is not an efficient choice for asynchronous FL with heterogeneous devices. 

In summary, we conclude that the joint design of scheduling and aggregation for asynchronous FL requires different considerations than the synchronous 	FL case. Furthermore, our proposed asynchronous FL scheme with periodic aggregation provides an efficient and flexible structure to resolve the straggler issue in synchronous FL systems.
	
\begin{table}[t!]
	\centering
	\caption{Legend description in Figs. \ref{fig:algo-100} and \ref{fig:algo-50}.}
	\renewcommand{\arraystretch}{1.2}
	\begin{tabular}{|c|p{1cm}|c|c|c|c|}
		\hline   
		Scheduling Policy & Legend &Aggregation Policy & Legend  \\
		\hline
		random  &rdm & age-aware, $\gamma=1.17$& fOld \\
		\hline
		significance-based& sgnfc   &age-aware, $\gamma=0.85$  &fFresh\\
		\hline
		frequency-based & freq &  &  \\
		\lasthline
	\end{tabular}
	\vspace{-.5cm}
	\label{tab:legend}
\end{table}	

\section{Conclusions}
In this work, we proposed an FL framework with asynchronous local training and periodic update aggregation. Specifically, we considered an asynchronous FL system over a resource-limited network where only a fraction of devices are allowed to upload their local model updates to the server in every communication round. Several device scheduling and update aggregation polices were investigated and compared through simulations. We observed that random scheduling performs surprisingly better than the alternative options for our proposed asynchronous FL scheme, especially under non-IID data distribution. Due to different levels of data freshness caused by asynchronous training, an appropriate age-aware model aggregation design can also greatly affect the system performance.

\end{document}